# BanglaNLP at BLP-2023 Task 1: Benchmarking different Transformer Models for Violence Inciting Text Detection in Bangla

**Saumajit Saha**
saha.saumajit@gmail.com

**Albert Nanda**
albert.nanda@gmail.com

## Abstract

This paper presents the system that we have developed while solving this shared task on violence inciting text detection in Bangla. We explain both the traditional and the recent approaches that we have used to make our models learn. Our proposed system helps to classify if the given text contains any threat. We studied the impact of data augmentation when there is a limited dataset available. Our quantitative results show that finetuning a multilingual-e5-base model performed the best in our task compared to other transformer-based architectures. We obtained a macro F1 of 68.11% in the test set and our performance in this shared task is ranked at 23 in the leaderboard.

## 1 Introduction

Violence inciting text detection (VITD) is the task of identifying text that incites violence in the Bangla language. This is a challenging task due to the complexity of the Bangla language and the variety of ways in which violence can be incited. The VITD task is important for several reasons. First, it can help to prevent violence by identifying and removing inciting text before it can cause harm. Second, it can help to protect people from being targeted by violence. Third, it can help to build a more peaceful and tolerant society.

There are several challenges to VITD in Bangla, and one of them is the scarcity of annotated data, which is due to the limited number of datasets of Bangla text that have been annotated for violence. This makes it difficult to train machine learning models that can accurately detect violence inciting text. Another challenge is the complexity of the Bangla language. Bangla is a morphologically rich language, which means that words can have multiple meanings depending on their context. This can make it difficult to identify violence inciting text, as the same words can be used in both violent and non-violent contexts. Despite these challenges, there has been some progress in the development of VITD systems for Bangla. The VITD task is still a research area, and there is still much work to be done.

## 2 Related Works

Sharif et al. (2022) introduced a multilabel dataset in Bangla to do aggressive text classification with a hierarchical annotation scheme. (Jahan et al., 2022) created a new Bangla Hate dataset and proposed BanglaHateBERT for abusive language detection in Bangla. (Romim et al., 2022) introduced a manually labeled large hate speech dataset in Bangla.

## 3 System Description

This section describes our system which is developed to classify violence inciting text written in Bangla. This section starts with the shared task description, followed by the description of the dataset released by the shared task organizers, then our proposed architecture which has produced our team's standing on the leaderboard and finally the results achieved and observations made. All the codes and datasets used for performing the experiments are available in https://github.com/Saumajit/BanglaNLP/tree/main/Task_1.

### 3.1 Shared Task Description

The objective of this shared task[1] (Saha et al., 2023a) is to identify the threats associated with violence in a given text segment. Given a Bangla text segment as input, the output produced by the system should belong to one of the 3 classes - *Non-Violence*, *Passive Violence* and *Direct Violence*.

### 3.2 Dataset Description

The dataset (Saha et al., 2023b) comprises YouTube comments related to the top 9 violent incidents that have occurred in the Bengal region

---
[1] https://github.com/blp-workshop/blp_task1

| Sentence | Label |
|---|---|
| একজন বাবা কতোটা অসহায় হলে এই কথা বলতে পারে আল্লাহ তুমি বিচার করো | Non-Violence |
| অসৎ এর বাচ্চারা তোরা কলেজ বিশ্ববিদ্যালয়ে এগুলো করার জন্যেই যাস, আর তোদের জন্য নীরিহ মানুষরা মারা যায় | Passive Violence |
| এই শালারে জন সম্মুখে আগুনে পুড়িয়ে মারা হউক, যাতে করে আর কোন অমানুষ এ রকম কাজ করতে সাহস না পায় । | Direct Violence |

Table 1: Sample dataset for each of the categories

| Actual Sentence | Augmented Sentence | Label |
|---|---|---|
| হিজাবেই নারীর সৌন্দর্য ফুটে ওঠে। | হিজাবে নারীর সৌন্দর্য প্রতিফলিত হয়। | Non-Violence |
| ভোট চুরি করলে তো খুন হবেই। | যদি আপনি ভোট চুরি করেন, তাহলে আপনাকে হত্যা করা হবে। | Passive Violence |
| সকল ছাত্রদের এক হয়া উচিত এবং নিউমার্কেট কে বয়কট করা উচিত। | সকল ছাত্র-ছাত্রীকে একত্রিত হতে হবে এবং নিউমার্কেট বর্জন করতে হবে। | Direct Violence |

Table 2: Comparison of actual and augmented data across different categories.

(Bangladesh and West Bengal) within the past 10 years. The dataset encompasses content in Bangla, with comment lengths of up to 600 words. *Non-Violence* refers to the category that pertains to non-violent subjects, such as discussions about social rights or general conversational topics that do not involve any form of violence. In *Passive Violence*, instances of violence are represented by the use of derogatory language, abusive remarks, or slang targeting individuals or communities. Additionally, any form of justification for violence is also classified under this category. *Direct Violence* refers to the category which encompasses explicit threats directed towards individuals or communities, including actions such as killing, rape, vandalism, deportation, desocialization (threats urging individuals or communities to abandon their religion, culture, or traditions), and resocialization (threats of forceful conversion). In Table 1, we can see a snippet of how the sentences in the dataset look like for each of the different categories. Table 3 highlights the distribution of different categories across train and development splits of the dataset.

| Class Labels | Train | Dev |
|---|---|---|
| Non-Violence | 1389 | 717 |
| Passive Violence | 922 | 417 |
| Direct Violence | 389 | 196 |

Table 3: Dataset distribution across train and development sets.

### 3.2.1 Data Augmentation

Finetuning deep learning models requires a lot of data for better performance on the desired task. However, we do not often have a large dataset available. We then require to increase the size of our dataset based on the limited dataset available to us. Feng et al. (2021) highlighted the different approaches available for doing data augmentation in NLP. From Table 3, we can understand that the amount of training data for every category is on the lower side. We therefore tried to augment data by using the Paraphrasing technique to generate text that will try to resemble actual data.

We used *bnaug*[2] library for augmenting data. We augmented 500 samples of each of the Non-Violence and Passive Violence categories. We augmented 389 samples of the Direct Violence category. We randomly chose samples from each category in the training set and then augmented one new sample for each original sample belonging to the training set. We had also tried to augment more number of samples for all categories to create a larger dataset but that led to inferior model performance. Table 2 shows a sample of augmented sentences corresponding to actual sentences for each of the categories.

### 3.3 Our Approaches

We performed several experiments to solve this task. We started with traditional machine learning algorithms like *Logistic Regression*, *Multinomial*

---
[2]https://github.com/sagorbrur/bnaug

*Naive Bayes* (Kibriya et al., 2005), *SGD Classifier*, *Majority Voting* (Lam and Suen, 1997) of earlier approaches and *Stacking* with *XGBoost* (Chen and Guestrin, 2016) as the final classifier. We used TF-IDF (Ramos, 2003) vectorization to convert words into vectors before feeding them to the machine learning algorithms. Table 4 highlights their performance on the development set. These experiments were performed on the actual data split provided, without doing any data augmentation.

| Algorithms | Macro-F1 |
|---|---|
| Logistic Regression | 52.97% |
| SGD Classifier | 44.8% |
| Multinomial Naive Bayes | 52.13% |
| Majority Voting of above three | 51.67% |
| Stacking | 50.99% |

Table 4: Performance of Traditional ML algorithms on the development set

Since we are solving a classification task where the contextual meaning of the sentence matters, we also experimented with several transformer (Vaswani et al., 2017) architectures to see how they perform in this task. We studied the impact data augmentation has when data are scarce and the model is unable to generalize well on unseen data. We used the AutoModelForSequenceClassification class from Hugging Face for finetuning all the models we discussed next.

We initially started with *BanglaBERT* (Sarker, 2020) which is nothing but base ELECTRA (Clark et al., 2020) model pre-trained with Replaced Token Detection objective. This model had been pretrained on the huge amount of web-crawled data and post-filtering to include only Bangla data. We finetuned BanglaBERT in this shared task's dataset using a learning rate of $5e-5$, batch size of 32, and number of epochs set to 4.

We then experimented with the multilingual version of Bert (Devlin et al., 2019), that is, *bert-base-multilingual*[3] which was pretrained using 104 languages. We used a learning rate of $5e-6$ and a batch size of 32, and the best model was obtained after finetuning for 3 epochs.

We also studied how the recently released and very popular multilingual models available in Hug-ging Face, *multilingual-e5-base*[4] and *multilingual-e5-large*[5] (Wang et al., 2022), perform in our task. Both these models were initialized from *xlm-roberta-base*[6] and *xlm-roberta-large*[7] respectively during pretraining. They undergo a two-stage training process - 1. Contrastive pretraining with unlabelled text pairs to gain a solid foundation on general-purpose embeddings, 2. Supervised training with labeled data so that human knowledge can be injected into the model and it is shown to boost performance. During our finetuning on the shared task's dataset, we used a learning rate of $5e-5$, batch size of 32, and number of epochs as 4. We also prepended a prompt (পাঠ্য অংশের অনুভূতি শ্রেণীবদ্ধ করুন:) to the input text during finetuning of both the variants of multilingual-e5.

[3]https://huggingface.co/bert-base-multilingual-cased

### 3.4 Results and Findings

The evaluation metric for this shared task is Macro-F1. Macro-F1 calculates F1 for each label and finds their unweighted mean. This does not take label imbalance into account. Table 5 highlights the results obtained for different finetuned models with and without applying data augmentation during the development phase. We observed that data augmentation positively impacted model performance, providing significant gains in macro-F1 score. We also found that *multilingual-e5-base* with data augmentation performed the best out of all the experiments performed for this task. We thus chose this finetuned model for inference on the test set and obtained a macro F1 of 68.11% in the test set released during the evaluation phase of this task.

### 3.5 Error Analysis on Test set

This subsection dives deep into the performance of the model. It provides an analysis of the correct and incorrect predictions of the model on the test set during the evaluation phase. Table 6 highlights a few examples across different categories where the model makes incorrect predictions.

We analyzed the sentences that had been misclassified for each category individually. We looked at the *n-grams* present in those sentences and demonstrated a few of the most commonly oc-curring *n-grams* in Figure 1. For example, when the ground truth is *Non-Violence*, Figure 1 shows that the presence of three phrases has confused the model to make the prediction incorrectly. Similarly, we also found examples of other phrases that may have confused the model for other labels.

[4]https://huggingface.co/intfloat/multilingual-e5-base
[5]https://huggingface.co/intfloat/multilingual-e5-large
[6]https://huggingface.co/xlm-roberta-base
[7]https://huggingface.co/xlm-roberta-large

| Models | Without Augmentation Macro-F1 | With Augmentation Macro-F1 | Change in Macro-F1 |
|---|---|---|---|
| BanglaBERT | 64.5% | 69.4% | +4.9 |
| Bert-base-multilingual-cased | 67.2% | 69.3% | +2.1 |
| Multilingual-e5-base | 71.57% | **74.6%** | +3.03 |
| Multilingual-e5-large | 60.48% | 69.36% | +8.88 |

Table 5: Finetuning-based experimental Results on the development set

| Sentence | Ground Truth | Prediction |
|---|---|---|
| দুই হাজার আটের পরে বাংলাদেশে কোনো বিচার হয়নি | Passive Violence | Non-Violence |
| ওরা রাজনিতি কে কাজে লাগায় বিচার হয় দলীয় ভাবে | Passive Violence | Non-Violence |
| এদের থেকে বাংলাদেশের নারীদের শিক্ষা নেওয়া উচিত | Non-Violence | Direct Violence |
| হিজাব বন্ধ করে ঠিকই করেছে। | Direct Violence | Non-Violence |
| হিন্দুদের কে ভাল করে সাইজ করা অচিত | Direct Violence | Non-Violence |
| ভারতের স্কুল খোলা অথচ বাংলাদেশে স্কুল বন্ধ। ভারতের মহামারী এখন বাংলাদেশে? | Non-Violence | Direct Violence |

Table 6: Snippet of incorrect prediction on the test set.

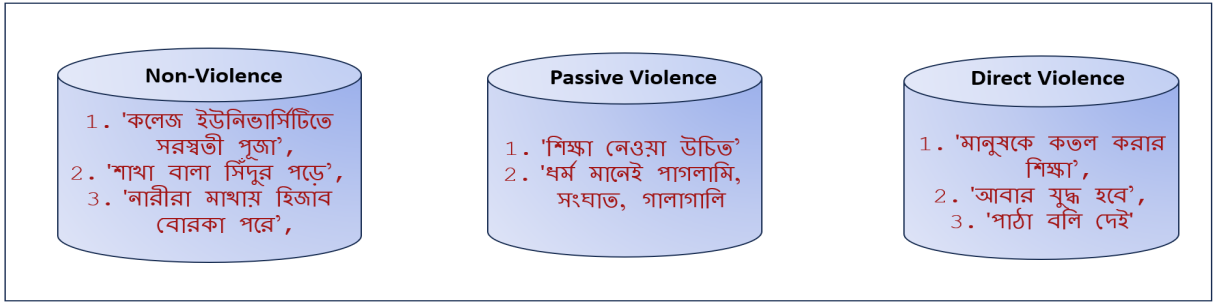

Figure 1: Snippet of phrases that the model has failed to capture correctly.

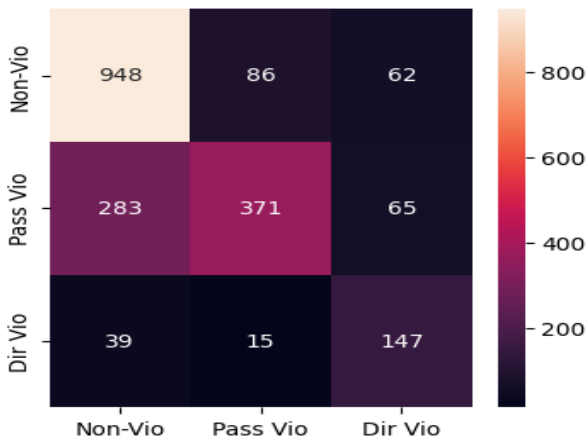

Figure 2: Confusion matrix on the test set. Non-Vio : Non-Violence, Pass Vio : Passive Violence, Dir Vio : Direct Violence.

Figure 2 highlights the confusion matrix our model's predictions produce on the test set. We observed that out of 201 sentences having *Direct Violence* as ground truth, 147 had been correctly predicted by the model. Similarly, 948 out of 1096 instances had been successfully predicted as *Non-Violence*, and 371 out of 719 instances had been correctly classified as *Passive Violence*. We therefore understand that our model is more accurate in understanding *Non-Violence* and *Direct Violence* categories and it needs to improve for *Passive Violence* category.

## 4 Conclusion

This paper reports the experiments we performed using the transformer-based models to solve this task. We show the impact that data augmentation has while dealing with smaller datasets. Future research direction can include exploring recently released large language models to solve similar tasks in a low-resource language like Bangla.

## 5 Limitations

The experiments reported in this paper have produced results in the particular setting of hyperparameters mentioned as well as in the dataset shared by the shared task organizer. We do not do exhaustive hyperparameter optimization for all the experiments reported because of compute constraints. We also do not use ChatGPT anywhere in our experimentation and data augmentation because of pricing constraints. All the experiments are run on Google Colab mostly using V100 and T4 GPUs.

## References

Tianqi Chen and Carlos Guestrin. 2016. Xgboost: A scalable tree boosting system. In *Proceedings of the 22nd acm sigkdd international conference on knowledge discovery and data mining*, pages 785–794.

Kevin Clark, Minh-Thang Luong, Quoc V. Le, and Christopher D. Manning. 2020. ELECTRA: pre-training text encoders as discriminators rather than generators. *CoRR*, abs/2003.10555.

Jacob Devlin, Ming-Wei Chang, Kenton Lee, and Kristina Toutanova. 2019. BERT: Pre-training of deep bidirectional transformers for language understanding. In *Proceedings of the 2019 Conference of the North American Chapter of the Association for Computational Linguistics: Human Language Technologies, Volume 1 (Long and Short Papers)*, pages 4171–4186, Minneapolis, Minnesota. Association for Computational Linguistics.

Steven Y. Feng, Varun Gangal, Jason Wei, Sarath Chandar, Soroush Vosoughi, Teruko Mitamura, and Eduard Hovy. 2021. A survey of data augmentation approaches for NLP. In *Findings of the Association for Computational Linguistics: ACL-IJCNLP 2021*, pages 968–988, Online. Association for Computational Linguistics.

Md Saroar Jahan, Mainul Haque, Nabil Arhab, and Mourad Oussalah. 2022. BanglaHateBERT: BERT for abusive language detection in Bengali. In *Proceedings of the Second International Workshop on Resources and Techniques for User Information in Abusive Language Analysis*, pages 8–15, Marseille, France. European Language Resources Association.

Ashraf M. Kibriya, Eibe Frank, Bernhard Pfahringer, and Geoffrey Holmes. 2005. Multinomial naive bayes for text categorization revisited. In *AI 2004: Advances in Artificial Intelligence*, pages 488–499, Berlin, Heidelberg. Springer Berlin Heidelberg.

L. Lam and S.Y. Suen. 1997. Application of majority voting to pattern recognition: an analysis of its behavior and performance. *IEEE Transactions on Systems, Man, and Cybernetics - Part A: Systems and Humans*, 27(5):553–568.

Juan Ramos. 2003. Using tf-idf to determine word relevance in document queries.

Nauros Romim, Mosahed Ahmed, Md Saiful Islam, Arnab Sen Sharma, Hriteshwar Talukder, and Mohammad Ruhul Amin. 2022. Bd-shs: A benchmark dataset for learning to detect online bangla hate speech in different social contexts. In *Proceedings of the Language Resources and Evaluation Conference*, pages 5153–5162, Marseille, France. European Language Resources Association.

Sourav Saha, Jahedul Alam Junaed, Maryam Saleki, Mohamed Rahouti, Nabeel Mohammed, and Mohammad Ruhul Amin. 2023a. Blp-2023 task 1: Violence inciting text detection (vitd). In *Proceedings of the 1st International Workshop on Bangla Language Processing (BLP-2023)*, Singapore. Association for Computational Linguistics.

Sourav Saha, Jahedul Alam Junaed, Maryam Saleki, Arnab Sen Sharma, Mohammad Rashidujjaman Rifat, Mohamed Rahout, Syed Ishtiaque Ahmed, Nabeel Mohammad, and Mohammad Ruhul Amin. 2023b. Vio-lens: A novel dataset of annotated social network posts leading to different forms of communal violence and its evaluation. In *Proceedings of the 1st International Workshop on Bangla Language Processing (BLP-2023)*, Singapore. Association for Computational Linguistics.

Sagor Sarker. 2020. Banglabert: Bengali mask language model for bengali language understanding.

Omar Sharif, Eftekhar Hossain, and Mohammed Moshiul Hoque. 2022. M-BAD: A multilabel dataset for detecting aggressive texts and their targets. In *Proceedings of the Workshop on Combating Online Hostile Posts in Regional Languages during Emergency Situations*, pages 75–85, Dublin, Ireland. Association for Computational Linguistics.

Ashish Vaswani, Noam Shazeer, Niki Parmar, Jakob Uszkoreit, Llion Jones, Aidan N. Gomez, Lukasz Kaiser, and Illia Polosukhin. 2017. Attention is all you need. *CoRR*, abs/1706.03762.

Liang Wang, Nan Yang, Xiaolong Huang, Binxing Jiao, Linjun Yang, Daxin Jiang, Rangan Majumder, and Furu Wei. 2022. Text embeddings by weakly-supervised contrastive pre-training.